\pdfoutput=1

\documentclass[11pt]{article}

\usepackage[]{ACL2023}

\usepackage{times}
\usepackage{latexsym}

\usepackage[T1]{fontenc}

\usepackage[utf8]{inputenc}

\usepackage{microtype}

\usepackage{inconsolata}

\title{Evaluating Distributed Representations for Multi-Level Lexical Semantics: A Research Proposal}

\author{Zhu Liu \\
  Tsinghua University \\
  Computational Linguistics, School of Humanities \\
  \texttt{liuzhu22@mails.tsinghua.edu} \\
  }

\usepackage{booktabs}
\usepackage{tabularx}
\usepackage{caption}
\usepackage{amsfonts}
\usepackage{amsmath}
\usepackage{color,xcolor} 
\usepackage{xspace}
\usepackage{adjustbox}
\usepackage{multirow}
\newcommand*{\eg}{e.g.\@\xspace}
\newcommand*{\ie}{i.e.\@\xspace}

\usepackage{amssymb}
\usepackage{algpseudocode}
\usepackage{xcolor,colortbl}
\usepackage{pifont}
\usepackage[utf8]{inputenc}
\usepackage{CJKutf8}
\usepackage[T5]{fontenc}

\makeatletter

\newcommand{\Rmnum}[1]{\mathrm{\expandafter\@slowromancap\romannumeral #1@}}
\makeatother

\begin{document}

\setcounter{page}{1}

\maketitle
\begin{abstract}

Modern neural networks (NNs), trained on extensive raw sentence data, construct distributed representations by compressing individual words into dense, continuous, high-dimensional vectors. These representations are expected to capture multi-level lexical meaning. In this thesis, our objective is to examine the efficacy of distributed representations from NNs in encoding lexical meaning. Initially, we identify and formalize three levels of lexical semantics: \textit{local}, \textit{global}, and \textit{mixed} levels. Then, for each level, we evaluate language models by collecting or constructing multilingual datasets, leveraging various language models, and employing linguistic analysis theories. This thesis builds a bridge between computational models and lexical semantics, aiming to complement each other.

\end{abstract}

\section{Introduction}
\label{introduction}


A key issue in lexical semantics is the many-to-many The relation between form and meaning is multifaceted. One form can encompass multiple senses, such as in cases of homonymy and polysemy. Conversely, one meaning can correspond to multiple forms, \eg, synonymy. Together, these phenomena organize a lexicon into a semantic field~\cite{jackson2000words} with varying levels~\cite{lietard-etal-2024-word}. 
From a local perspective, each word possesses potential meanings influenced by different contexts. These meanings may be unrelated (homonymy) or related, with varying degrees of relatedness. At a global level, words are interconnected through relationships such as word analogy (\textit{man} $-$ \textit{woman} $=$ \textit{king} $-$ \textit{queen}) and compositionality (\textit{Italy} $+$ \textit{noodles} $=$ \textit{spaghetti}). 
At a mixed level, we can conceive an overarching conceptual space that captures cross-lingual universality, integrating both words and their local meanings.


Modern computational language models based on transformer architectures~\cite{vaswani2017attention} represent a word~\footnote{The actual representation for a model corresponds to a token, \ie, subword. We can represent a word by aggregating the contextual embeddings from subwords in practice.} as a real-valued, contextual, high-dimensional vector~\cite{break-petersen-potts-2023-lexical} (or representation). This approach is grounded in the distributional hypothesis~\cite{harris1954distributional}, which posits that words appearing in similar contexts have similar meanings. Vectors from pre-trained language models (PLMs) and large language models (LLMs) have been utilized as initial inputs for downstream tasks related to lexical semantics, achieving outstanding performance in areas such as word sense disambiguation~\cite{bevilacqua2021recent}, lexicon induction~\cite{li2023bilingual}, and reverse dictionary tasks~\cite{tian2024prompt}.
Natural and important questions arise: how well do these distributed representations from PLMs and LLMs convey contextual meaning? Do they truly possess the same lexical knowledge as humans, given their impressive linguistic performance? If not, to what extent do they fall short? Answering these questions is significant for both the computational and linguistic communities. For computational researchers, these insights could enhance the transparency and reliability of black-box models. For linguists, they could aid in constructing meaning systems and discovering novel meanings.



This thesis explores these research questions by evaluating current pre-trained language models (PLMs) and large language models (LLMs) on a variety of lexical meaning tasks with differing levels of granularity. To incorporate a cross-lingual perspective, we design benchmarks spanning multiple languages. Furthermore, we test our hypotheses by integrating theories and models from both computational linguistics and traditional linguistic frameworks.

\paragraph{Research Proposal}
In this thesis, we first formalize our research problems, including key concepts and the three levels of analysis (see Section~\ref{sec:formalization}). Next, we collect datasets and design experiments to evaluate each aspect (see Section~\ref{sec:evaluation}). Finally, we address potential methodological challenges and present the conclusions drawn from our findings (see Section~\ref{sec:conclusion}).

\section{Formalization}
\label{sec:formalization}

\subsection{Basic Notions}
We consider a lexicon $\mathcal{W}$ consisting of a finite and countable set of items. These items include all lemmatized words $w_i \in \mathcal{W}$, each associated with a prototypical meaning $m(w_i)$. This prototypical meaning represents the most frequent and typical sense recognized by speakers of a given language community~\cite{rosch1975cognitive}, and is typically listed first in a dictionary. For instance, the prototypical meaning of $m_{\text{bank}}$ refers to a financial institution, rather than the side of a river or a shape of fog. These prototypical meanings form a meaning space $\mathcal{M}$. 

Words are then sampled and combined into sequences, forming a corpus $\mathcal{R}$, consisting of sentences $s_i$. For a word $w_i$, its occurrence $o(w'_i)$ is realized as a sentence $s_i$ within the corpus $\mathcal{R}$, where $w'_i$ represents a conjugated form of $w_i$, such as the addition of inflectional morphemes\footnote{This process occurs in the surface structure derived from the deep structure, according to generative grammar~\cite{chomsky1965aspects}.}. The word $w_i'$ in its context $s_i$ carries a conventional meaning, referred to as its sense $e(w'_i)$, which is typically listed in other dictionary entries, distinct from the prototypical meaning $m(w_i)$. These senses, across lexical items, constitute a conceptual field $\mathcal{C}$, where each concept represents a distinct type of meaning. Thus, a concept is the minimal unit through which we understand the world.
There are several examples of the combination of $\mathcal{W}$, $\mathcal{R}$, and $\mathcal{C}$.

\paragraph{Example 1} A common dictionary for learners, such as the Oxford Dictionary~\footnote{\url{https://en.oxforddictionaries.com/}}, includes words $w$ with their associated meanings $m$, and sentences derived from a corpus $\mathcal{R}$. While the entire conceptual space is implicit, it is suggested through sets of synonyms.

\paragraph{Example 2} WordNet~\cite{WordNet} contains words, their meanings, and sentences, similar to a conventional dictionary. Additionally, synsets in WordNet, as nodes, explicitly represent a conceptual space by grouping synonymous words.

\paragraph{Example 3} The mental lexicon~\cite{klepousniotou2002processing} is inherently present in our minds. Words in the vocabulary are learned incrementally from fragments of utterances in $\mathcal{R}$. The conceptual space helps us recognize equivalent concepts and facilitate categorization.

\begin{figure}
    \centering
    \includegraphics[width=0.9\linewidth]{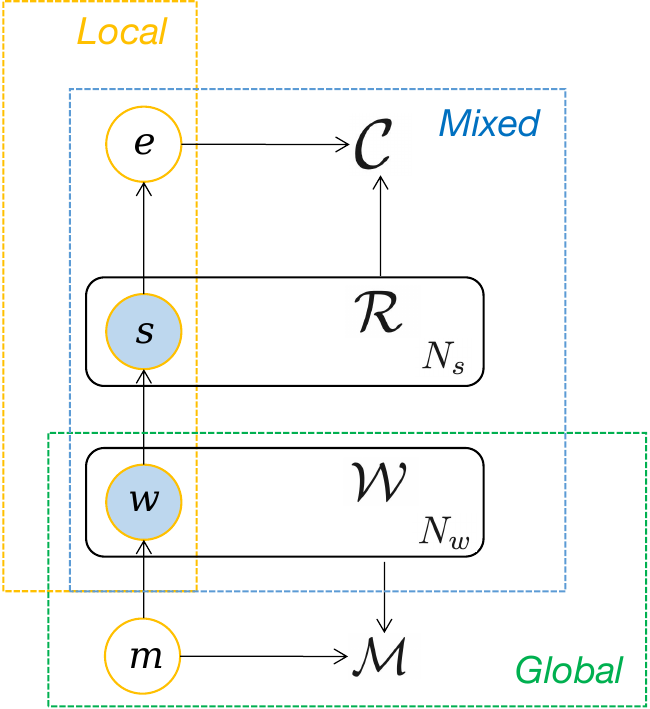}
    \caption{Graph models for four spaces: lexicon $\mathcal{W}$, corpus $\mathcal{R}$, concept $\mathcal{C}$, and prototypical meaning $\mathcal{M}$. The three levels and their respective scopes are identified. The shaded circle represents an observable variable, while the unshaded one indicates an unobservable variable.}
    \label{fig:graph_model}
\end{figure}

\subsection{Three Levels}
\begin{figure}
    \centering
    \includegraphics[width=1.0\linewidth]{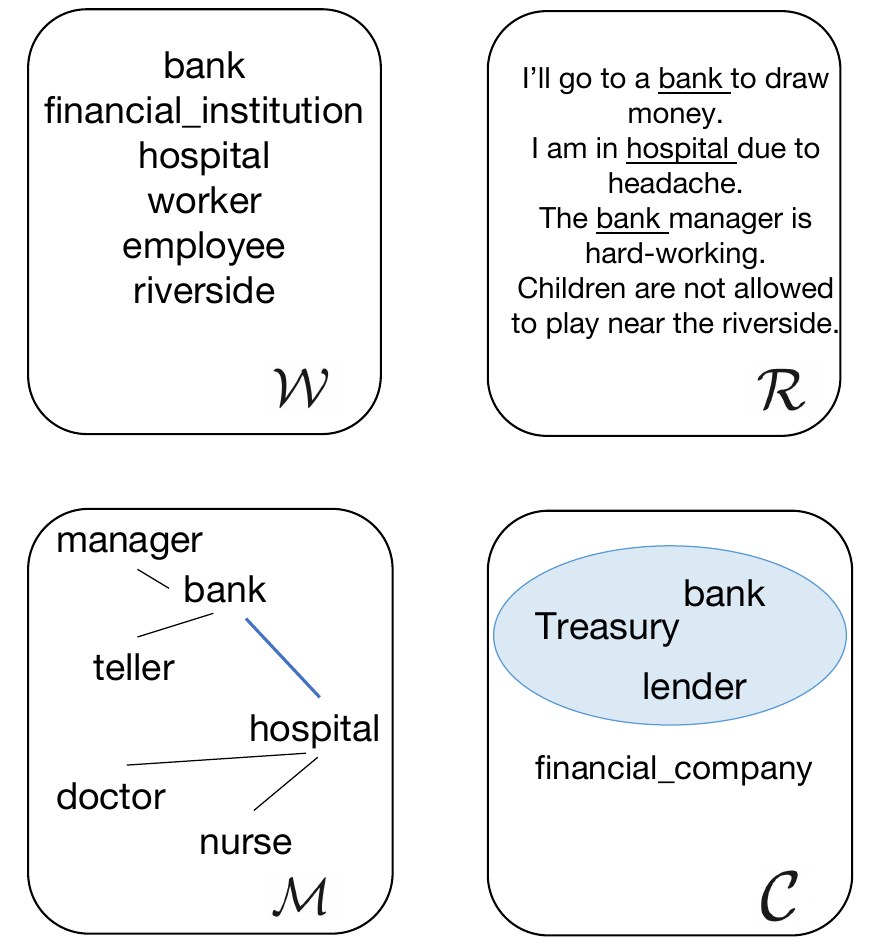}
    \caption{Examples of the four spaces: lexicon $\mathcal{W}$, corpus $\mathcal{R}$, concept $\mathcal{C}$, and prototypical meaning $\mathcal{M}$.}
    \label{fig:four_spaces}
\end{figure}

We propose a graphical model to illustrate the relationships among four spaces: $\mathcal{W}$, $\mathcal{M}$, $\mathcal{R}$, and $\mathcal{C}$, as shown in Figure~\ref{fig:graph_model}. Further examples are provided in Figure~\ref{fig:four_spaces}. Based on these, we identify three hierarchical levels.

\paragraph{Local Level} This level focuses on the sense space within a word, addressing issues such as the number of senses, how senses are divided and evolve, the similarity and relatedness of potential senses, as well as conventional or temporal meanings. Tasks like word sense disambiguation~\cite{2009survey}, word sense induction~\cite{van-de-cruys-apidianaki-2011-latent}, and lexical semantic changes~\cite{schlechtweg2023human} reflect the local space of a word. We formalize this level using the following likelihood probability:
\begin{equation}
    p(e) = p(e|w,s),
\end{equation}
where $w \in \mathcal{W}$, $s \in \mathcal{R}$, and $e \in \mathcal{C}$.

\paragraph{Global Level} From another perspective, we consider all the words in a lexicon and examine their semantic relationships. Since no specific context is provided at this level, we focus solely on the prototypical meaning of words. Relations among words include word analogy~\cite{word2vec}, word similarity~\cite{huang2012improving}, compositionality~\cite{ica_yamagiwa-etal-2023-discovering}, and others. Structures such as vectors and graphs are commonly used to represent these relationships. We formalize this level using the following likelihood probability:
\begin{equation}
    p(\mathcal{M}) = p([e_i]_N | \mathcal{W}),
\end{equation}
where $N$ is the number of words in the lexicon, and $[e_i]_N = \{e_1, e_2, \dots, e_i, \dots, e_N\}$.
\paragraph{Mixed Level}
The mixed level considers the senses across all words in the lexicon. A semantic field can form a set of concepts shared within a language community. Building on the local level, problems such as concept induction and the similarity of concepts are classical challenges at this level. A conceptual space~\cite{haspelmath2003geometry} that discovers relations among concepts, often in a cross-lingual context, represents the structure of concepts. The process at the mixed level can be realized in two steps: first, local-level sense induction, followed by global concept induction. We formalize this level using the following likelihood probability:
\begin{equation}
\small
    p(\mathcal{C}) = p([c_i]_M |\mathcal{W}, \mathcal{R}) = p(e | w,s) \cdot p([c_i]_M|e, \mathcal{W}),
\end{equation}
where $M$ is the number of concepts, and $[c_i]_M = \{c_1, c_2, \dots, c_i, \dots, c_M\}$.

In this paper, we evaluate to what extent distributed representations convey lexical semantics at these three levels. At the local level (Subsection~\ref{subsection: local}), we focus on the phenomenon of continuous relatedness of senses and investigate whether language models can reflect the continuous variation in different senses. At the global level (Subsection~\ref{subsection: global}), we construct a network of word embeddings learned by large language models and analyze their geometric relations among different words. For the mixed level (Section~\ref{subsection: mixed}), we build conceptual spaces from a subset of the lexicon based on typological theories in language.

\section{Evaluation}
\label{sec:evaluation}

We begin by utilizing various types of language models to extract representations, followed by an evaluation of these representations across the three levels.

\subsection{Models and Representations}
\label{subsec:representation}
Drawing on the Distributional Hypothesis~\cite{harris1954distributional}, language models employ neural networks to derive continuous vectors from large-scale corpora. Models such as Word2Vec~\cite{word2vec} and GloVe~\cite{pennington-etal-2014-glove} generate static word representations that do not account for the different senses a word may take in varying contexts. In contrast, transformer-based models~\cite{kenton2019bert} acquire layer-wise contextual representations. Both static and contextual representations leverage the distribution of neighboring words but do not explicitly model meaning in a nuanced manner. Some studies, however, aim to build sense representations in an unsupervised~\cite{liu2015learning} or knowledge-based~\cite{chen2014unified} fashion.

To evaluate the representations from these models, we consider different configurations, including BERT-like bidirectional models (primarily used in pre-trained language models, PLMs) and GPT-like generative models (commonly employed in large language models, LLMs). We propose to investigate how LLMs encode lexical semantics~\cite{llm2vec,liu-etal-2024-fantastic} in comparison to the more established research on BERT (BERTology~\cite{rogers2021primer}). Specifically, we aim to identify where lexical semantics are encoded within the model, which will guide how we extract vectors for further evaluation.

Additionally, we propose transforming the original semantic space into a more discrete form to decouple interdependent features. Techniques such as independent component analysis~\cite{ica_yamagiwa-etal-2023-discovering} can be employed for this purpose. By analyzing the values of the transformed feature axes, we seek to identify meaningful dimensions of semantics that better capture the underlying structure of lexical meaning.

\subsection{Local Level: Sense Continuity}
\label{subsection: local}
At the local level, we focus on the distribution of possible senses for words. A well-known phenomenon in lexical semantics is the continuity of sense distinctions. That is, the relatedness of senses for different words varies continuously. Linguists distinguish between homonymy and polysemy, for example. However, even within polysemy, the distance between senses for some words in different contexts can vary. Furthermore, variations in semantic roles for words within the same context can also differ. These challenges are reflected in the high annotator disagreement found in lexical semantics-related tasks, such as word sense disambiguation (WSD)~\cite{2009survey}.

\paragraph{Uncertainty in WSD}
Existing supervised methods~\cite{bevilacqua2021recent} treat WSD as a classification task and have achieved remarkable performance. However, they often overlook uncertainty estimation (UE) in real-world settings, where sense labeling involves substantial disagreement and varying degrees of uncertainty for different words~\cite{liu-liu-2023-ambiguity}. We propose addressing this issue by formalizing it as a probabilistic inference problem~\cite{gal2016uncertainty}, which outputs a well-calibrated probability distribution over the candidate sense space.

Incorporating uncertainty estimation also helps differentiate between homonymy and polysemy. Homonymy tends to exhibit higher certainty in its sense labeling, whereas polysemy is often characterized by greater uncertainty. This perspective aligns with the inherent subjectivity of language, which includes phenomena such as underspecification, vagueness, and context sensitivity~\cite{sep-ambiguity}.



\paragraph{Semantic Roles}
\label{subsection:SR}
In a basic SVO sentence structure based on predicate-argument relationships, the degree of semantic roles assigned to different grammatical constituents (S, V, and O) can vary. This phenomenon is linguistically universal. For instance, the verb \textit{shot at} in the sentence \textit{The hunter shot at the bear} exhibits weaker transitivity than \textit{shot} in \textit{The hunter shot the bear.} In Chinese, word order plays a significant role in determining semantic roles; typically, the first argument is the agent, and the second is the patient. Changing the subject and object order can alter the agency and objectivity, although in some cases, the semantic roles remain unchanged. For example, \begin{CJK}{UTF8}{gbsn} 张三打李四 \end{CJK} (Tom hit Alice) is fundamentally different from \begin{CJK}{UTF8}{gbsn} 李四打张三 \end{CJK} (Alice hit Tom). However, \begin{CJK}{UTF8}{gbsn} 十个人吃一顿饭 \end{CJK} (Ten people eat a meal) essentially carries the same meaning as \begin{CJK}{UTF8}{gbsn} 一顿饭吃十个人 \end{CJK} (A meal provides ten people to eat). In the former case, the verb \textit{hit} has the same transitivity in both contexts, while \textit{eat} does not. We propose collecting minimal pairs (where only the subject and object change) that represent different shifts in semantic roles and investigating whether language models can capture such nuances. The forms of minimal pairs may vary across different languages because distinct languages may employ different strategies, such as morphological or syntactic variations, to reflect changes in semantic roles. We emphasize that the form of the minimal pair is specific to each language, as languages may have distinct ways of modifying the degree of semantic roles.

This work is related to the classical NLP task of semantic role labeling~\cite{jurafskyspeech}, but with several key differences. First, we treat the degree of semantic roles as a continuous variable, rather than a binary choice. Thus, we utilize representations to calculate the similarity between corresponding items. Second, we collect minimal pairs where only one factor changes, such as the transitivity of verbs, to facilitate causal analysis. Finally, we consider cross-lingual universality and compare the behavior of semantic role assignment across different languages.

\subsection{Global Level: Word Network}
\label{subsection: global}
At the global level, we propose constructing a word network by leveraging the embeddings learned in large language models (LLMs). In this network, nodes represent words, and edges reflect the similarity between corresponding words. This structure illustrates how the internal representations of models capture the relationships among words. We also aim to compare networks built using models of varying scales. Several factors need to be addressed. First, how should we extract the representation of a word from raw \textbf{token} embeddings? Does the conventional mean-pooling approach remain effective? Second, what similarity measure should be employed? While cosine similarity is commonly used, it may not suffice in higher-dimensional spaces. Third, how should we prune a fully connected graph, particularly when comparing different models? What constitutes a fair strategy for pruning across different model types? Finally, how can we design effective indicators to zoom in and out of the network? When zooming in~\cite{li2024geometry}, we can observe detailed examples such as meaning analogies (e.g., man-king-woman-queen). Conversely, zooming out allows us to characterize the entire network through graph statistics, such as the number of connected components.

\subsection{Mixed Level: Conceptual Spaces}
\label{subsection: mixed}
At the mixed level, we consider the senses across words and construct a conceptual space that reflects the similarity of concepts. Given the cross-lingual universality of concepts, we adopt the theory of language typology, specifically semantic map modeling~\cite{haspelmath2003geometry} (SMM), where a conceptual space is built based on a connectivity hypothesis.

Function words, affixes, and certain adverbs play a crucial role in SMM due to their multifunctionality. They exhibit a broader range of nuanced semantics or functions compared to content words and are often not exhaustively listed in dictionaries. For example, repetitive grams (such as "and" or "again") in various languages can demonstrate over 20 distinct functions~\cite{zhang2017semantic}. Linguists use Semantic Map Models (SMM)~\cite{haspelmath2003geometry} to visually represent these functions within a conceptual or semantic space, interconnected by lines to form a network. Functions with greater similarity are positioned closer together on the map. SMM is grounded in cross-linguistic comparison, following the "semantic connectivity hypothesis," which suggests that functions expressed by a language-specific category should occupy contiguous areas on the semantic map. 

In our approach, we utilize representations from language models to measure the similarity between different occurrences of a target word. Subsequently, we design a graph algorithm to construct the semantic map, adhering to the connectivity principle. We aim to assess the quality of the automatically generated graph against a human-annotated one using designated metrics.
·

\section{Conclusion and Challenges}
\label{sec:conclusion}
Distributed representations encode rich lexical semantics, capturing not only word meanings but also contextual associations, allowing for a nuanced understanding of language. This thesis evaluates the extent to which vectorized representations reflect word meaning across three levels: the local level (word-relatedness), the global level (multilingual lexicon relations), and the mixed level (cross-lingual conceptual space). These levels offer both micro and macro perspectives of semantic fields, providing a comprehensive framework for evaluating lexical semantics. By considering these multiple levels, we aim to better understand how models capture the complexity of language. We propose evaluating model representations using common benchmarks and custom datasets, with performance serving as an indicator of semantic quality and the ability to reflect the depth of meaning encoded in these representations.

However, several challenges arise in this probing approach. The first is the ``extraction dilemma'': can poor task performance be attributed to a model’s failure to capture semantics, or does it result from suboptimal representation extraction strategies? Without isolating other factors, we cannot conclusively assess the semantic capacity of the representations. The second issue is probe selectivity~\cite{hewitt-liang-2019-designing}, where it’s unclear if the probe extracts representations or simply learns the task. The third challenge is dataset bias, as contextual meanings are often more subjective than static meanings. Disagreements among human annotators on tasks like polysemy disambiguation can introduce uncertainty, affecting metric design and reliability.

Lastly, while scaling laws~\cite{kaplan2020scaling} show that larger models improve performance, they also increase opacity, raising questions about model interpretability and trustworthiness. How do these models arrive at their conclusions? Can we explain their predictions in a way that is meaningful to humans? Our research aims to enhance the transparency of modern language models and bridge the gap between computer science and linguistics, fostering a better understanding of how these models represent linguistic information.



\bibliography{custom}
\bibliographystyle{acl_natbib}




\end{document}